\documentclass[lettersize,journal]{IEEEtran}
\usepackage{amsmath,amsfonts}
\usepackage{algorithmic}
\usepackage{algorithm}
\usepackage{array}
\usepackage[caption=false,font=normalsize,labelfont=sf,textfont=sf]{subfig}
\usepackage{textcomp}
\usepackage{stfloats}
\usepackage{url}
\usepackage{verbatim}
\usepackage{graphicx}
\usepackage{tikz}
\usepackage{pgfplots} %引用包
\usepackage{cite}
\usepackage{multirow}
\usepackage{xcolor}

\usepackage{scalerel}
\usepackage{tikz}
\usetikzlibrary{svg.path}
\definecolor{orcidlogocol}{HTML}{A6CE39}
\tikzset{
    orcidlogo/.pic={
        \fill[orcidlogocol] svg{M256,128c0,70.7-57.3,128-128,128C57.3,256,0,198.7,0,128C0,57.3,57.3,0,128,0C198.7,0,256,57.3,256,128z};
        \fill[white] svg{M86.3,186.2H70.9V79.1h15.4v48.4V186.2z}
        svg{M108.9,79.1h41.6c39.6,0,57,28.3,57,53.6c0,27.5-21.5,53.6-56.8,53.6h-41.8V79.1z M124.3,172.4h24.5c34.9,0,42.9-26.5,42.9-39.7c0-21.5-13.7-39.7-43.7-39.7h-23.7V172.4z}
        svg{M88.7,56.8c0,5.5-4.5,10.1-10.1,10.1c-5.6,0-10.1-4.6-10.1-10.1c0-5.6,4.5-10.1,10.1-10.1C84.2,46.7,88.7,51.3,88.7,56.8z};
    }
}
\newcommand\orcidicon[1]{\href{https://orcid.org/#1}{\mbox{\scalerel*{
                \begin{tikzpicture}[yscale=-1,transform shape]
                \pic{orcidlogo};
                \end{tikzpicture}
            }{|}}}}
\usepackage[implicit=false]{hyperref}
\hypersetup{hidelinks,
	colorlinks=true,
	allcolors=black,
	pdfstartview=Fit,
	breaklinks=true}
\begin{document}

\title{Intra-task Mutual Attention based Vision\\ Transformer for Few-Shot Learning}

\author{Weihao Jiang, Chang Liu, Kun He${\textsuperscript{\orcidicon{0000-0001-7627-4604}}}$,~\IEEEmembership{Senior Member,~IEEE}
        % <-this % stops a space
%\thanks{This paper was produced by the IEEE Publication Technology Group. They are in Piscataway, NJ.}% <-this % stops a space

\thanks{Weihao Jiang, Chang Liu and Kun He are with the School of Computer Science, Huazhong University of Science and Technology, Wuhan, China 430074. Corresponding author: Kun He, E-mail: brooklet60@hust.edu.cn.}
\thanks{}
\thanks{This work was supported by National Natural Science Foundation (U22B2017).}
}

\markboth{}%
{Jiang \MakeLowercase{\textit{et al.}}:Intra-task Mutual Attention based Vision Transformer for Few-Shot Learning}

\IEEEpubid{}

\maketitle

\begin{abstract}
Humans possess remarkable ability to accurately classify new, unseen images after being exposed to only a few examples. Such ability stems from their capacity to identify common features shared between new and previously seen images while disregarding distractions such as background variations. However, for artificial neural network models, determining the most relevant features for distinguishing between two images with limited samples presents a challenge. In this paper, we propose an intra-task mutual attention method for few-shot learning, that involves splitting the support and query samples into patches and encoding them using the pre-trained Vision Transformer (ViT) architecture. 
Specifically, we swap the class (CLS) token and patch tokens between the support and query sets to have the mutual attention, which enables each set to focus on the most useful information. 
This facilitates the strengthening of intra-class representations
and promotes closer proximity between instances of the same class. For implementation, we adopt the ViT-based network architecture and utilize pre-trained model parameters obtained through self-supervision. By leveraging Masked Image Modeling as a self-supervised training task for
pre-training, the pre-trained model yields semantically meaningful representations while successfully avoiding supervision collapse. We then employ a meta-learning method to fine-tune %the last two 
the last several layers and CLS token modules. Our strategy significantly reduces the number of parameters that require fine-tuning while effectively utilizing the capability of pre-trained model. Extensive experiments show that our framework is simple, effective and computationally efficient, achieving superior performance as compared to the state-of-the-art baselines on five popular few-shot classification benchmarks under the 5-shot and 1-shot scenarios.
\end{abstract}


\section{Introduction}
% \IEEEPARstart{T}{his} file is intended to serve as a ``sample article file''

%With the boom of deep learning,
%In recent years, 
There has been a substantial body of researches in the realm of few-shot learning (FSL)~\cite{siamese,one-shot,Auto-FSL,cgf}, whose goal is to imitate human's proficiency of acquiring new concepts, even with a small number of provided training instances. 
FSL models aim to learn from a minimal set of examples and effectively generalize the learned knowledge to diverse and previously unseen images. 
Researches of FSL can be categorized into two main types. The first is meta-learning, which utilizes samples from a base class and optimizes the model for target new class, and metric-based approaches, that primarily involves classifying images based on the metric values between support and query features, stand as the most widely adopted methods.
The second is transfer learning, where a feature extractor is trained on a rich base class which is subsequently used to predict weights of the new class classifier. 
%In the realm of meta-learning, metric-based approaches currently stand as the most widely adopted methods. This methodology primarily involves classifying images based on the metric values between support and query features. 
% To enhance the representation of images, scholars have sequentially delved into methods based on local feature maps, global features, and the utilization of both.
%Among these, feature extraction methods can be broadly categorized into two main types: those based on Convolutional Neural Networks (CNNs) and those based on Vision Transformer (ViT) models.
All these FSL methods involve a key component of feature extraction, using a backbone of Convolutional Neural Networks (CNNs) and recently Vision Transformer (ViT) models.

% The CNN-based methods primarily utilizes local feature maps as the representation of images. To garner more distinctive features, researchers~\cite{CTM,CAN,CTX,ATL-Net,RENet} introduce external modules to generator weight mask to enhance the features. These methods share a common limitation, namely, their reliance on local features. This constraint may stem from the convolution operation employed by CNN in image processing, which makes CNN-based method relatively less adept at processing global information in images. 
Early FSL methods are based on CNN models~\cite{gli,msml,qsformer,RGTransformer,CTM,CAN,CTX,ATL-Net,RENet}. 
These CNN-based methods primarily utilize local feature maps as the representation of images. To obtain more distinctive features, researchers~\cite{CTM,CAN,CTX,ATL-Net,RENet} introduce external modules to generate weight masks to enhance these features. However, these methods are collectively limited by their dependence on local features.
This limitation may arise from the convolution operation employed by CNN in image processing, making CNN-based methods relatively less adept at processing global information in images.

\IEEEpubidadjcol
\begingroup
\renewcommand{\arraystretch}{1.5}
\begin{table*}[t]
\caption{Comparison between our proposed IMAFormer and the counterparts. In the pursuit of acquiring more discriminative features, CNN-based models frequently incorporate an additional module, leveraging local feature maps to identify crucial local regions for image representation. On the other hand, while ViT-based models inherently possess the capability to capture both global and local features, existing methods often utilize them in isolation. In contrast, our approach adopts the ViT architecture and innovatively integrates global and local features through our intra-task mutual attention method, aiming to derive more discriminative features.}
\begin{center} 
\begin{tabular}{l|c|c|c}
\hline 
Model  & Feature utilization method     & Backbone   & External module \\  
\hline 
CTM~\cite{CTM}, CAN~\cite{CAN}, CTX~\cite{CTX}, ATL-Net~\cite{ATL-Net}, ReNet~\cite{RENet}   & Only operate local feature map    &ResNet & Yes    \\ 
\hline
PMF~\cite{P>M>F}    &    Only use class token    &     \multirow{3}{*}{ViT}                 & No              \\
FewTURE~\cite{FewTrue}   &Only use patch tokens  &           & Yes             \\
HCTransformer~\cite{HCTransformer}     &     Independent use of class token and patch tokens  &   & Yes  \\ 
\hline 
%\hline
%Integrated use between 
 IMAformer (ours)  & Cross-use
class token and patch tokens              & ViT                    & No              \\ \hline
\end{tabular}
\end{center}
\label{tab:comparation}
\end{table*}
\endgroup

More recently, there is a growing body of FSL works based on Vision Transformers~\cite{P>M>F,FewTrue,HCTransformer}. 
Approaches based on ViT facilitates the acquisition of both global feature (class token) and local feature (patch token) of an image by partitioning it into multiple non-overlapping patches and leveraging the attention mechanism. While ViT excels in extracting global features, current methods predominantly employ it as an efficient feature extractor, falling into three main categories.
The first method~\cite{P>M>F} involves extracting class tokens from support set images and query images using a pre-trained model, then the similarity between these class tokens serves as a metric for evaluation. The second method~\cite{FewTrue} focuses on extracting patch tokens from the images and learning a weight generator based on the patch tokens of the support set to enhance the feature representation. 
The third method~\cite{HCTransformer} utilizes both class token and patch tokens through the introduction of a potential supervised propagation scheme. However, it is noteworthy that despite combining these two features, they are utilized independently in the method.

% More recently, Vision Transformers (ViT)~\cite{ViT,BEiT,MAE,DINO} have emerged as a promising alternative to CNNs in various computer vision tasks, showcasing substantial potential. By segmenting input samples into patches and leveraging Vision Transformers for encoding, we can establish both global features (class token) and local features (patch token). Despite ViT's %unique 
% advantages in extracting global features, current approaches primarily utilize it as a potent feature extractor.
% Focusing on few-shot learning, methods based on Vision Transformers predominantly address three key aspects. Firstly, the model~\cite{P>M>F} is trained on base class data using a pre-trained model, followed by fine-tuning all parameters using the support set from test data. This method employs the model to extract class tokens from support set images and the query image, measuring their similarity based on these class tokens. Secondly, FewTURE~\cite{FewTrue} extracts patch tokens of the image and learns a weight generator based on the local features of the support set to enhance feature representation. Thirdly, HCTransformer~\cite{HCTransformer} introduces a latent supervision propagation scheme based on both class and patch tokens, although the two types of features are employed independently.

% While these methods have demonstrated notable performance improvements compared to CNN models, they still fall short of fully integrating global and local features to attain more discriminative representations.
ViT-based approaches demonstrate significant performance improvements compared to the CNN-based models. However, they still face the challenge of adequately considering the support set and query set across the entire task. Despite the capability of ViT-based approaches to capture both local and global features of the images, they generate features independently for the support set and the query set. Consequently, in the generation of global features for the support set, the query set's local features has not been taken into account. Similarly, the generation of global features for the query set overlooks the contribution of local features from the support set. This leads to the inability to obtain a more discriminative representation for the task.

We observe that humans carefully examine known support images one by one to identify similar features to the query, particularly the prominent local features. In this way, humans can grasp both global information and local information, as well as make two-way comparisons between the query and support. 
Thereby, we introduce 
an intra-task mutual attention based on Vision Transformer, called IMAFormer, where the query and support sets 
pay attention to each other 
and reinforce common features of the same category. 

Diverging from previous approaches, our method leverages a fusion of both global and local features from support and query sets, without resorting to external modules beyond the backbone architecture, as shown in Table \ref{tab:comparation}. 
Specifically, we divide the support and query sets into patches and utilize the first $L-1$ layers of the ViT model to obtain the CLS token and patch tokens. These patch tokens are then exchanged between the query and support sets. Through the attention mechanism, the query set pays attention to the detailed features of the support set, absorbing and strengthening shared characteristics between the same class. Likewise, the support set focuses on what the query concerns, reinforcing features of the same category as the query within the support set. Such process increases the similarity within the same category between the query and support sets while decreasing the similarity between different categories.

Our method employs a Transformer architecture, incorporates the concept of intra-task mutual attention, utilizes the meta-learning method to fine-tune specific parameters based on the self-supervised learning pre-trained model training with Masked Image Modelling as pretext task, and  leads to boosted results in few-shot learning tasks. Our main contributions are summarized as follows:

\begin{itemize}
\item %To our best knowledge, we innovatively propose, for the first time, 
We propose a novel method of simultaneous utilizing both global and local features from support and query data to enhance the features of valuable targets for few-shot learning based on Vision Transformers.
\item We propose a novel few-shot learning framework that enables the model to learn more discriminative features by incorporating an intra-task mutual attention module to the Vision Transformer architecture.
\item Our method is simple, efficient and effective, based on self-supervised learning pre-trained models, only need to fine-tune fewer parameters, and can outperform existing methods without external module. 
\item We provide the efficacy of our method by achieving new state-of-the-art results on five popular public benchmarks, and do ablation study to show the effectiveness of our intra-task mutual attention method.
\end{itemize}

\section{Related Works}

In this section, we first summarize the categories of few-shot learning methods and analyze the distinctions between our approach and existing methods. Next, we introduce several ViT pre-training methods and pick Masked Image Modeling as our pre-training approach. Finally, we present the application of cross-attention, emphasizing the differences in our intra-task mutual attention.

\subsection{Few-shot Learning}

In recent years, there has been a significant body of research dedicated to the few-shot learning task. Broadly, these efforts can be categorized into two main branches: meta-learning and transfer learning.
Certainly, most of the existing few-shot learning methods rely on the meta-learning framework, which can be roughly divided into two categories, optimization-based and metric-based. 

\textbf{Optimization-based methods in FSL.} Optimization-based methods aim to quickly learn the model parameters that can be optimized when encountering new tasks. As a representative work of this category, MAML~\cite{MAML} obtains good initialization parameters through internal and external training in the meta-training stage. 
MAML has inspired many follow-up efforts, such as ANIL~\cite{ANIL}, BOIL~\cite{BOIL}, and LEO~\cite{LEO}.

\textbf{Metric-based methods in FSL.} Metric-based methods embed support images and query images into the same space and classify query images by calculating the distance or similarity. 
Using different metric methods, various models have been derived, such as Matching Networks~\cite{MatchingNetwork}, Prototypical Networks~\cite{PrototypicalNetwork}, MSML~\cite{msml} and DeepEMD~\cite{deepemd}. Simultaneously, to garner more distinctive features, researchers have started treating the entire task holistically and devising external modules to enhance the features.
CTM~\cite{CTM} introduces a Category Traversal Module that traverses the entire support set at once, discerning task-relevant features based on both intra-class commonality and inter-class uniqueness in the feature space. CAN~\cite{CAN} generates cross-attention maps for each pair of support feature map and query sample feature map to emphasize target object regions, enhancing the discriminative nature of the extracted features. CTX~\cite{CTX} proposes learning spatial and semantic alignment between CNN-extracted query and support features using a Transformer-style attention mechanism. ATL-Net~\cite{ATL-Net} introduces episodic attention calculated by a local relation map between the query image and the support set, adaptively selecting important local patches within the entire task. RENet~\cite{RENet} combines self-correlational representation within each image and cross-correlational attention modules between images to learn relational embeddings.

Recently, researchers have started incorporating ViT into few-shot learning scenarios. 
The PMF~\cite{P>M>F} method consists of three phases. Firstly, it utilizes a pre-trained Transformer model with external unsupervised data. Then, it simulates few-shot tasks for meta-training using base categories. Finally, the model is fine-tuned using scarcely labeled data from test tasks.
HCTransformer~\cite{HCTransformer} employs hierarchically cascaded Transformers as a robust meta feature extractor for few-shot learning.
It utilizes the DINO~\cite{DINO} learning framework and maximizes the use of annotated data for training. Additionally, it enhances data efficiency through attribute surrogates learning and spectral tokens pooling.
FewTURE~\cite{FewTrue} adopts a completely transformer-based architecture and learns a token importance weight through online optimization during inference.

% Relation Networks calculate distances between support images and query images via a relation module. TADAM~\cite{fc100} boosts Prototypical Networks by metric scaling and metric task conditioning. DeepEMD~\cite{deepemd} employs the Earth Mover’s Distance (EMD) as a metric to compute a structural distance between dense support images and query images.

\textbf{FSL methods based on transfer learning.} Few-shot learning methods based on the transfer learning framework use a simple scheme to train a classification model on the overall training set, and yields comparable results. During testing,
it removes the classification head and retains the feature extraction part, then trains a new classifier based on the support set from testing data.  The representative jobs are Dynamic Classifier~\cite{dynamic}, Baseline++~\cite{baseline++} and RFS~\cite{RFS}.

Our proposed method falls into the category of metric-based meta-learning methods. 
In contrast to existing metric-based learning approaches that extract support and query sample features separately, our method exploits the semantic correlation between support and query features to emphasize the target object. 
While some methods~\cite{CTM,CAN,CTX,ATL-Net,RENet} also consider the relationship between support and query samples, they rely on local feature maps and complex external module networks. 
In contrast, our approach achieves superior performance without the need for additional computational overhead, utilizing self-supervised pre-trained models. 
As demonstrated in the experimental section, our method outperforms these approaches by a significant margin.

\subsection{Self-supervised Pretraining ViT}
Vision Transformers have shown superior performance compared to traditional architectures~\cite{dsp,HRTransNet}. However, their effectiveness heavily relies on a vast amount of labeled data, which may not always be feasible or sustainable for supervised training. Thus, implementing a self-supervised approach for Vision Transformers can make them powerful and more applicable to downstream problems.
BEiT~\cite{BEiT} introduces the concept of Masked Image Modeling as a self-supervised training task for training ViT.
%During pre-training, BEiT tokenizes the original image, randomly masks image blocks, and feeds the masked image into the encoder. The main goal of pre-training is to restore the masked image block based on the unmasked counterpart.
SimMIM~\cite{SimMIM} suggests that randomly masking image parts and increasing the resolution of image blocks can yield satisfactory results.
%It proposes that the optimal performance is achieved when the image block resolution is 32$\times$32.
In contrast to BEiT, MAE~\cite{MAE} simplifies the training logic by using random masks to process input image blocks and directly reconstructing the masked image blocks for training. 
%MAE incorporates two key designs. First, it employs an encoder-decoder architecture with an asymmetric structure, where the encoder only considers non-masked image blocks, along with a lightweight decoder design. Second, it covers a majority of the image blocks, such as using a mask probability of 75\%, to obtain a more meaningful self-supervised training task.
DINO~\cite{DINO} presents a straightforward self-supervised approach for ViT, which can be viewed as a form of knowledge distillation without the use of labels. 
The DINO framework involves predicting the output of a teacher network, constructed with a momentum encoder, by employing a standard cross-entropy loss.

In our approach, we leverage a Vision Transformer architecture as an encoder, chosen for its patch-based nature. Simultaneously, to address the issue of supervision collapse, we implement self-supervised training, employing Masked Image Modeling as a pretext task. Moreover, a considerable number of Transformer models pre-trained using self-supervised methods exist. To expediently assess the extensibility of our model, we incorporated certain pre-trained model parameters when utilizing the ViT-Base model, specifically those obtained through pre-training with the MAE method.

\subsection{Cross Attention}
Cross-attention was first introduced in the field of natural language processing~\cite{attention}, and the attention mechanism used to process sequence data applications in the field of computer vision. For image processing, convolutional neural network is usually used to extract the local features of the images. However, in order to better capture the global relationship, cross-attention introduces a mechanism that allows the model to allocate different attention when processing different parts of the input. CTX~\cite{CTX} proposes learning spatial and semantic alignment between CNN-extracted query
and support features using a cross attention mechanism. CrossViT~\cite{crossvit} propose a dual-branch transformer to combine image patches through token fusion module based on cross attention, which uses a single token for each branch as a query to exchange information with other branches. 

%Diverging from 
Different from conventional approaches, we engage in the interchange of global features between the support image and the query image. %Subsequently,
Specifically, we leverage the attention mechanism inherent in the Transformer encoder layer to facilitate the cross-attention. This obviates the necessity for a distinct cross-attention module, thereby mitigating the need to alter the model structure. 

\section{Methodology}
%We begin by presenting the foundational concepts of few-shot learning. Subsequently, we outline the framework of our proposed method. Finally, we provide a comprehensive explanation on the functionality of our intratask cross attention module.

%\subsection{Problem Definition}

This section outlines the framework of our proposed IMAformer model, and provides a comprehensive study on our intra-task mutual attention method.

\subsection{The Framework}

\begin{figure}[!t]
\begin{center}
%\fbox{\rule{0pt}{1in} \rule{.9\linewidth}{0pt}}
 % \includegraphics[width=0.9\linewidth]{svrg-3-2.svg}
  \includegraphics[width=1.0\linewidth]{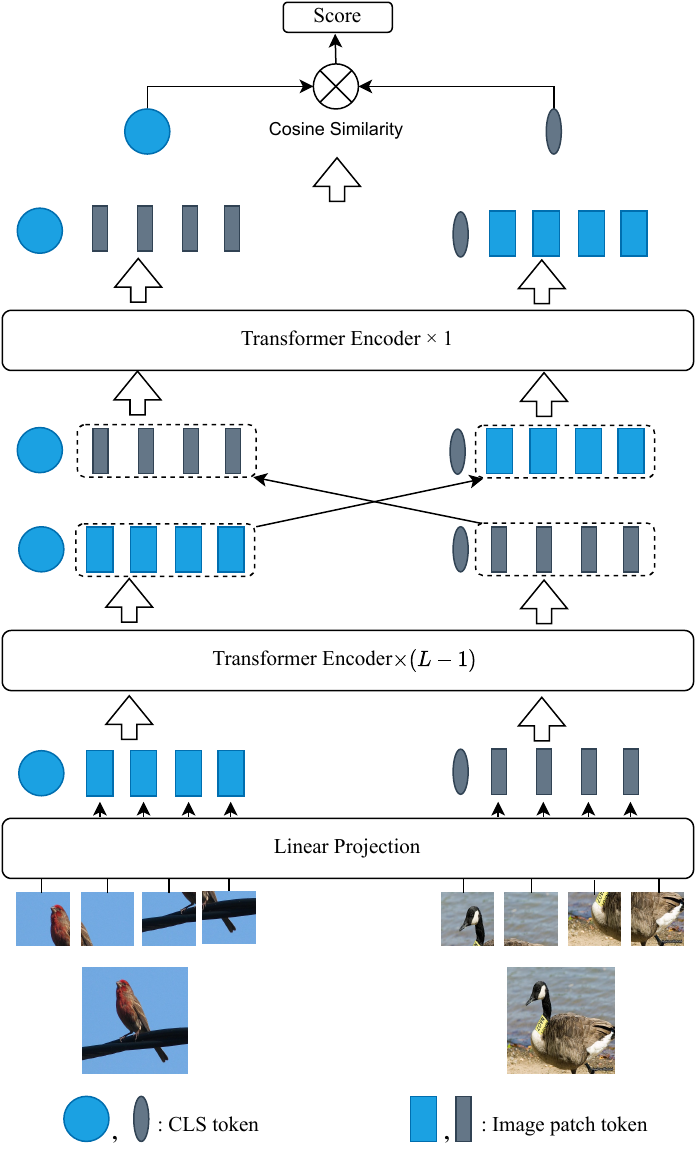}
\end{center}
   \caption{The main framework of our IMAformer model. We adopt a Transformer architecture with the first $L-1$ layers be used to encode the patched input images containing support and query. After obtaining the CLS token and patch tokens, we exchange the patch tokens between support and query, and the new combined tokens are sent to the last layer to get the final CLS token for the inputs.}
\label{fig:framework}
\end{figure}

The framework of our IMAformer model, as illustrated in Figure~\ref{fig:framework}, adopts a Transformer architecture. Initially, the input images are divided into non-overlapping patches. These patches, from both the support set and the query set, are then encoded using the first $L-1$ layers of ViT~\cite{ViT}, resulting in CLS tokens and patch tokens. The CLS tokens capture global semantic information of the entire image, whereas the patch tokens represent local information specific to each patch.
Then, we swap the patch tokens between the support image and the query image, such that the CLS tokens and patch tokens are crossed for the support set and query set. %merging them together. 
This allows the CLS token of the support image to be combined with the patch tokens of the query image, and vice versa. The merged tokens are then fed into the final layer of the ViT model, where attention mechanisms are employed to select relevant patches. 
In the end, we obtain the final CLS tokens for both the support images and the query images, for which we calculate the cosine similarity score to determine their similarity.  We can then assign the query image an  appropriate category basing on this score.

% 2023-12-21 add
Diverging from conventional approaches, we engage in the interchange of global features between the support image and the query image.  Subsequently, we leverage the attention mechanism inherent in the Transformer encoder layer to facilitate the cross-attention.  This obviates the necessity for a distinct cross-attention module, thereby mitigating the need to alter the model structure.
Our intra-task mutual attention method is parameter-free and can be seamlessly integrated into the existing Transformer Encoder without the need for architectural modifications or additional networks. It can be directly plugged in, enhancing the capabilities of the model without introducing extra parameters or complexity.

\subsection{Intra-task Mutual Attention}
In typical single few-shot task, there are generally five classes of support set. If there are $K$ shots in each category of support, take the average of the token values as the prototype tokens. After encoding, five sets of CLS token and patch token combinations can be obtained, namely:

\begin{figure}[!t]
\begin{center}
%\fbox{\rule{0pt}{1in} \rule{.9\linewidth}{0pt}}
 % \includegraphics[width=0.9\linewidth]{svrg-3-2.svg}
  \includegraphics[width=1.0\linewidth]{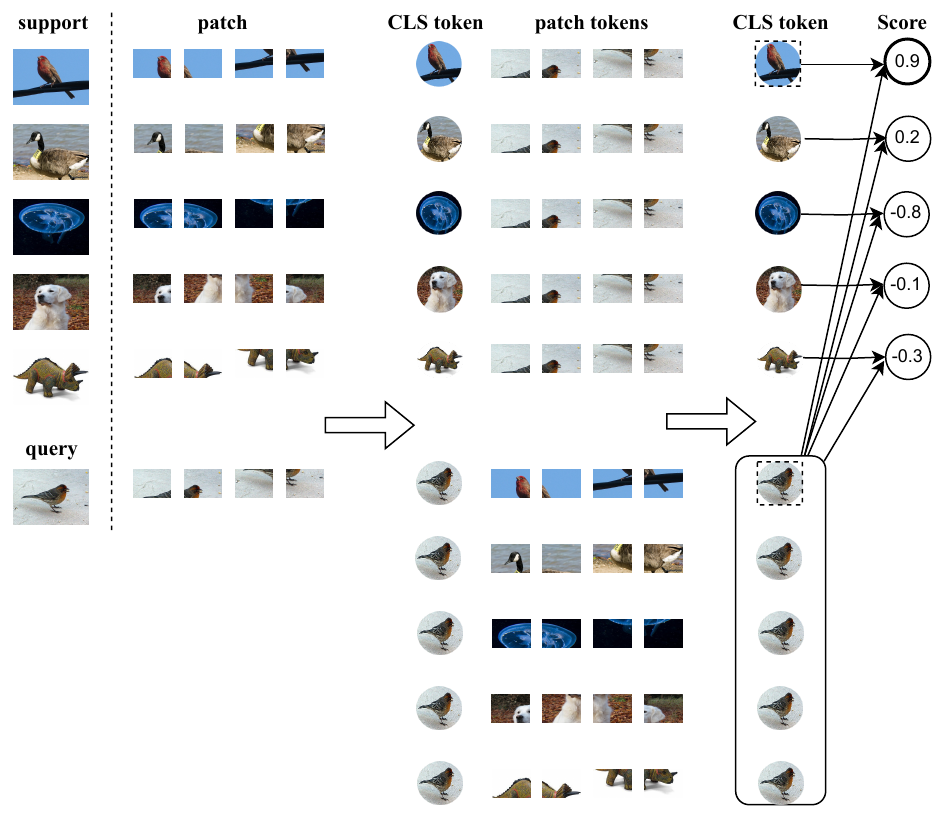}
\end{center}
   \caption{%Architecture of  the proposed IMAformer 
   Illustration of the IMAformer processing pipeline for 5-way 1-shot task. The support and query images are firstly patched and then the first $L-1$ layers of IMAformer are used to encode the input images. After getting the CLS token and patch tokens, the patch tokens between support and query are exchanged, and the new combination of tokens are sent to the last layer to get the final CLS token of input images. The CLS token of the same class between support and query will be strengthened, as marked by the dotted box.}
\label{fig:images}
\end{figure}

\begin{align}\label{e1}
    P_{i} &= [CLS_{i}^p,patch_{i,1}^p,patch_{i,2}^p,...,patch_{i,M}^p],\\
    &i\in [1,5], \nonumber
\end{align}
% $i\in [1,5]$,
where $M$ is the number of patches.

For a query image, we have a combination of CLS token and patch tokens:
\begin{align}\label{e2}
    Q  = [CLS^q,patch_{1}^q,patch_{2}^q,...,patch_{M}^q],
\end{align}
where $M$ is the number of patches.

To ensure that the query pays attention to the specific details of all categories in the support set, and to make the support set aware of the details of the query, we combine the CLS token of the query with the patch tokens of each category in the support set. Similarly, we combine the CLS tokens of each category in the support set with the patch tokens of the query. Such process is visualized in Figure~\ref{fig:images}. 

For $i\in [1,5]$, the combined tokens
\begin{align}\label{e3}
    P_{i}{}' &= [CLS_{i}^p,patch_{1}^q,patch_{2}^q,...,patch_{M}^q], 
    %&i\in [1,5]\nonumber
\end{align}

\begin{align}\label{e4}
    Q_{i}{}'  &= [CLS^q,patch_{i,1}^p,patch_{i,2}^p,...,patch_{i,M}^p]
    %&i\in [1,5]\nonumber 
\end{align}
are sent into the final layer, and the final CLS token is extracted as the representation for both the support and query: 
\begin{align}\label{e5}
    CLS_{p,i} = FFN(Softmax\left(\frac{W_q(CLS_{i}^p) \cdot W_k(P_{i}{}')}{\sqrt{d}}\right) \\
    \cdot W_v(P_{i}{}')),\nonumber 
\end{align}
\begin{align}\label{e6}
    CLS_{q,i} = FFN(Softmax\left(\frac{W_q(CLS^q) \cdot W_k(Q_{i}{}')}{\sqrt{d}}\right) \\
    \cdot W_v(Q_{i}{}')),\nonumber 
\end{align}
where $W_q$, $W_k$ and $W_v$ are weights of the fully connected layer, and $FFN$ is the feed forward network. 

By performing the above operations, each category in the support set gains the ability to perceive information of the query image, thereby reinforcing the representation if the support and query images are of the same class. Similarly, the query image obtains distinct feature representations by observing the feature details of each category in the support set separately. This process enhances the representations that are similar to the support set, resulting in more discriminative extracted features.
To calculate the similarity between the query image and the support set, the following approach can be employed:
\begin{align}\label{e7}
    cos(q,p_j) = \sum_{i=1}^{5}cos( CLS_{q,i},CLS_{p,j}). 
\end{align}

By strengthening the representation of the same class, the margin between the similarity of instances within the same class and the similarity between instances of different classes increases. It in turn leads to improved accuracy in classifying queries.
 
\subsection{Classification using Enhanced CLS token}

% \begin{align}\label{e8}
%     \mathbf{c}_n = \frac{1}{|S_n|} \sum_{(\mathbf{x},y) \in {S_n}}f_{\theta}(\mathbf{x}).
% \end{align}

Similarly, a query point $\mathbf{x}$ is passed through the feature extractor to obtain the corresponding feature representation. The cosine similarity between $\mathbf{x}$ and the support set prototypes is calculated to obtain the probability that $\mathbf{x}$ belongs to class $n$:
\begin{align}\label{e9}
    p(y=n|\mathbf{x}) = \frac{exp(cos(q_\mathbf{x},p_n ))}{\sum_{n=1}^{N}exp(cos(q_\mathbf{x},p_n))},
\end{align}
where $cos(\cdot,\cdot)$ denotes the cosine similarity between two vectors. 
In the end, the cross entropy $\mathcal{L}$ is used as the loss function.

% For a given task, the loss function can be formulated as follows:
% \begin{align}\label{e10}
%     L = -\frac{1}{NQ}\sum_{i=1}^{NQ}\sum_{i=1}^{N}I(y_i=n)\log p(y_i=n|x_i). 
% \end{align}

% \usepackage{multirow}
\begin{table*}[ht]
\caption{Comparison with the state-of-the-art 5-way 1-shot and 5-way 5-shot performance with 95$\%$ confidence intervals on $mini$Imagenet and $tiered$Imagenet. %In our methodology, we fine-tune the parameters of the last six layers when utilizing ViT-Small as the backbone, and fine-tune the parameters of the last two layers along with CLS token when ViT-Base is employed as the backbone. The highest-ranking results are emphasized in bold.
} 
\label{tab:miniImage}
\begin{center}
% \scalebox{1.0}{
    \begin{tabular}{l|cr|cc|cc}
    \hline
    \multirow{2}{*}{Model} &\multirow{2}{*}{Backbone} & \multirow{2}{*}{$\approx$Params} & \multicolumn{2}{c|}{$mini$ImageNet} & \multicolumn{2}{c}{$tiered$ImageNet} \\
                           &                  &              &1-shot& 5-shot & 1-shot& 5-shot \\ 
    \hline
    \hline
    ProtoNet~\cite{PrototypicalNetwork}  & ResNet-12 & 12.4 M & 62.29±0.33  & 79.46±0.48 & 68.25±0.23  & 84.01±0.56   \\
    FEAT~\cite{FEAT} & ResNet-12 & 12.4 M &66.78±0.20 &82.05±0.14 &70.80±0.23 &84.79±0.16 \\
    CAN~\cite{CAN} & ResNet-12 & 12.4 M& 63.85±0.48  & 79.44±0.34 &69.89±0.51 &84.23±0.37 \\
    CTM~\cite{CTM} & ResNet-18 & 11.2 M& 64.12±0.82 & 80.51±0.13 &68.41±0.39 &84.28±1.73 \\
    ReNet~\cite{RENet} & ResNet-12 & 12.4 M& 67.60±0.44  & 82.58 ±0.30 &71.61±0.51 &85.28±0.35  \\
    DeepEMD~\cite{deepemd} & ResNet-12 &12.4 M &65.91±0.82 &82.41±0.56 &71.16±0.87 &86.03±0.58 \\
    IEPT~\cite{IEPT} &ResNet-12 &12.4 M & 67.05±0.44 & 82.90±0.30 & 72.24±0.50 & 86.73±0.34 \\
    MELR~\cite{MELR} &ResNet-12 &12.4 M & 67.40±0.43 & 83.40±0.28 & 72.14±0.51 & 87.01±0.35 \\
    FRN~\cite{FRN} &ResNet-12 &12.4 M & 66.45±0.19 & 82.83±0.13 & 72.06±0.22 & 86.89±0.14 \\
    CG~\cite{CG/CNL} &ResNet-12 &12.4 M & 67.02±0.20 & 82.32±0.14 & 71.66±0.23 & 85.50±0.15 \\
    DMF~\cite{DMF} &ResNet-12 &12.4 M & 67.76±0.46 & 82.71±0.31 & 71.89±0.52 & 85.96±0.35 \\
    InfoPatch~\cite{InfoPatch} &ResNet-12 &12.4 M & 67.67±0.45 & 82.44±0.31 & - & -\\
    BML~\cite{BML} &ResNet-12 &12.4 M & 67.04±0.63 & 83.63±0.29 & 68.99±0.50 & 85.49±0.34 \\
    CNL~\cite{CG/CNL} &ResNet-12 &12.4 M & 67.96±0.98 & 83.36±0.51 & 73.42±0.95 & 87.72±0.75 \\
    Meta-NVG~\cite{Mata-NVG} &ResNet-12 & 12.4 M & 67.14±0.80 & 83.82±0.51 & 74.58±0.88 & 86.73±0.61 \\ 
    PAL~\cite{PAL} &ResNet-12 & 12.4 M & 69.37±0.64 & 84.40±0.44 & 72.25±0.72 & 86.95±0.47 \\
    COSOC~\cite{COSOC} &ResNet-12 & 12.4 M & 69.28±0.49 & 85.16±0.42 & 73.57±0.43 & 87.57±0.10 \\
    Meta DeepBDC~\cite{Meta-DeepBDC} &ResNet-12 & 12.4 M & 67.34±0.43 & 84.46±0.28 & 72.34±0.49 & 87.31±0.32 \\
    QSFormer~\cite{qsformer} &ResNet-12 & 12.4 M & 65.24±0.28 & 79.96±0.20 & 72.47±0.31 & 85.43±0.22 \\
    \hline
    LEO~\cite{LEO} & WRN-28-10 & 36.5 M & 61.76±0.08 & 77.59±0.12 & 66.33±0.05 & 81.44±0.09 \\
    CC+rot~\cite{CC-rot} & WRN-28-10 & 36.5 M & 62.93±0.45 & 79.87±0.33 & 70.53±0.51 & 84.98±0.36 \\
    FEAT~\cite{FEAT} & WRN-28-10 & 36.5 M & 65.10±0.20 & 81.11±0.14 & 70.41±0.23 & 84.38±0.16 \\
    % PSST~\cite{PSST} & WRN-28-10 & 36.5 M & 64.16±0.44 & 80.64±0.32 & - & - \\
    MetaQDA~\cite{MetaQDA} & WRN-28-10 & 36.5 M & 67.83±0.64 & 84.28±0.69 & 74.33±0.65 & 89.56±0.79 \\ 
    OM~\cite{OM} & WRN-28-10 & 36.5 M & 66.78±0.30 & 85.29±0.41 & 71.54±0.29 & 87.79±0.46 \\
    \hline
    FewTURE~\cite{FewTrue}  & ViT-Small & 22 M & 68.02±0.88 &  84.51±0.53 & 72.96±0.92 &  86.43±0.67\\
    FewTURE~\cite{FewTrue}  & Swin-Tiny & 29 M & 72.40±0.78 &  86.38±0.49 & 76.32±0.87 & 89.96±0.55\\
    HCTransformer\cite{HCTransformer} & ViT-Small & 22 M & 74.74±0.17  &  85.66±0.10  & 79.67±0.20  &  89.27±0.13 \\
    \hline
    % IMAformer (ours) & ViT-Small & 22 M & \emph{70.25±0.61}  &  \emph{86.48±0.44}  & \emph{75.15±0.67}  &  \emph{88.44±0.65} \\    % two layers       
    % IMAformer (ours) & ViT-Small & 22 M & \emph{78.48±0.39}  &  \emph{89.05±0.28}  & \emph{77.70±0.67}  &  \emph{90.68±0.65} \\
    IMAformer (ours) & ViT-Small & 22 M & 78.48±0.31  & 88.18±0.29  & 79.78±0.30  & 91.01±0.27 \\
    % \hline
    % \hline
    IMAformer (ours) & ViT-Base & 86 M & \bf85.68±0.66  &  \bf93.28±0.42  & \bf83.10±0.57  &  \bf93.10±0.55 \\
    \hline
    \end{tabular}
\end{center}

\end{table*}

\begin{table*}[ht!]
\caption{Comparison with the state-of-the-art 5-way 1-shot and 5-way 5-shot performance with 95$\%$ confidence intervals on CIFAR-FS and FC100.
%ViT-Base is the backbone, with parameters of the last two layers plus the CLS token be fine-tuned. At each column, the top two results are highlighted in bold and italic.}% based on their relative rankings.
}
\label{tab:cifar}
\begin{center}
% \scalebox{1.0}{
    \begin{tabular}{l|cr|cc|cc}
    \hline
    % \cline{1-1} \cline{3-5}
    \multirow{2}{*}{Model} &\multirow{2}{*}{Backbone} & \multirow{2}{*}{$\approx$Params} & \multicolumn{2}{c|}{CIFAR-FS} & \multicolumn{2}{c}{FC100} \\
             &          &            & 1-shot           & 5-shot         & 1-shot & 5-shot      \\ 
    \hline
    \hline
    ProtoNet~\cite{PrototypicalNetwork} & ResNet-12 & 12.4 M & - & -  &41.54±0.76 & 57.08±0.76 \\
    MetaOpt~\cite{MetaOpt} & ResNet-12 & 12.4 M & 72.00±0.70 & 84.20±0.50 & 41.10±0.60 & 55.50±0.60 \\
    MABAS~\cite{MABAS} & ResNet-12 & 12.4 M & 73.51±0.92 & 85.65±0.65 & 42.31±0.75 & 58.16±0.78 \\
    RFS~\cite{RFS} & ResNet-12 & 12.4 M & 73.90±0.80 & 86.90±0.50 & 44.60±0.70 & 60.90±0.60 \\
    BML~\cite{BML} & ResNet-12 & 12.4 M & 73.45±0.47 & 88.04±0.33 & -& - \\
    CG~\cite{CG/CNL} & ResNet-12 & 12.4 M & 73.00±0.70 &85.80±0.50 & -& - \\
    Meta-NVG~\cite{Mata-NVG} & ResNet-12 & 12.4 M & 74.63±0.91 & 86.45±0.59 & 46.40±0.81 & 61.33±0.71 \\ 
    RENet~\cite{RENet} & ResNet-12 & 12.4 M & 74.51±0.46 & 86.60±0.32 & - & - \\
    TPMN~\cite{TPMN} & ResNet-12 & 12.4 M & 75.50±0.90 & 87.20±0.60 & 46.93±0.71 & 63.26±0.74 \\
    MixFSL~\cite{MixFSL} & ResNet-12 & 12.4 M & -& -& 44.89±0.63 & 60.70±0.60 \\
    QSFormer~\cite{qsformer} &ResNet-12 & 12.4 M & - & - & 46.51±0.26 & 61.58±0.25 \\
    \hline
    CC+rot~\cite{CC-rot} & WRN-28-10 & 36.5 M & 73.62±0.31 & 86.05±0.22 & -& - \\
    PSST~\cite{PSST} & WRN-28-10 & 36.5 M & 77.02±0.38 & 88.45±0.35 & -& - \\
    Meta-QDA~\cite{MetaQDA} & WRN-28-10 & 36.5 M & 75.83±0.88 & 88.79±0.75 & - & -\\
    \hline
    FewTURE~\cite{FewTrue}  & ViT-Small & 22 M & 76.10±0.88 & 86.14±0.64 &46.20±0.79 &63.14±0.73 \\
    FewTURE~\cite{FewTrue}  & Swin-Tiny & 29 M & 77.76±0.81 &88.90±0.59 &47.68±0.78 &63.81±0.75 \\
    HCTransformer~\cite{HCTransformer} & ViT-Small & 22 M & 78.89±0.18 &87.73±0.11   & \bf48.27±0.15 & 61.49±0.15  \\
    \hline
    % IMAformer (ours) & ViT-Small & 22 M & \emph{78.57±0.73}	& \emph{88.04±0.63}	& \emph{46.38±0.53}	& \bf65.58±0.65 \\ two layers
    IMAformer (ours) & ViT-Small & 22 M & \bf81.46±0.73	& \bf92.06±0.63	& 46.38±0.53	& \bf65.58±0.65 \\
    % \hline
    % \hline
    IMAformer (ours) & ViT-Base & 86 M & 81.34±0.67	& 91.30±0.83 & 48.02±0.63	& 65.52±0.75 \\
    \hline
    \end{tabular}
\end{center}

\end{table*}

\section{Experiments}
 This section presents the experimental results, where we assess the characteristics of our model and compare the performance against several baseline methods on five widely used benchmark datasets. 
We also conduct an ablation study to demonstrate the effectiveness of our proposed method.

\subsection{Datasets}
In the standard few-shot classification task, there are five popular benchmark datasets for few-shot classification, including $mini$ImageNet~\cite{MiniImageNet}, $tiered$ImageNet~\cite{tieredImageNet}, CIFAR-FS~\cite{cifar-fs}, FC100~\cite{fc100} and CUB-200-2011~\cite{cub}.

% The $mini$ImageNet and the $tiered$ImageNet are both from ILSVRC-2012~\cite{ImageNet} and contain 100 and 608 categories, respectively.
% Both CIFAR-FS and FC100 are derived from CIFAR-100, which comprises 100 classes and 600 images per class. Unlike the previous datasets, CIFAR-FS and FC100 are characterized by small-resolution images, with dimensions of 32 × 32 pixels. 
% The CUB-200-2011 (CUB) dataset is for fine-grained classification and consists of 200 classes with a total of 11,788 images.

%The $mini$ImageNet dataset %was created by the authors of Matching Network~\cite{MatchingNetwork} and 
%is currently the most popular benchmark for FSL, which 
The $mini$ImageNet dataset contains 100 categories sampled from ILSVRC-2012~\cite{ImageNet}. Each category contains 600 images, and 60,000 images in total. In standard setting, it is randomly partitioned into training, validation and testing sets, each containing 64, 16 and 20 categories.

%The $tiered$ImageNet dataset is another commonly used benchmark. Although it is also from ILSVRC-2012, compared with the former, %the image size is the same, and 
%it has a more extensive data scale. 
The $tiered$ImageNet dataset is also from ILSVRC-2012, but with a more extensive data scale. 
There are 34 super-categories, partitioned into training, validation and testing sets, with 20, 6 and 8 super-categories, respectively. It contains 608 categories, corresponding to 351, 97 and 160 categories in each partitioned dataset. %This dataset is more challenging because it is more difficult for the model to identify samples of different categories from the same super-categories. The base classes and novel classes come from different super-categories. 

Both CIFAR-FS and FC100 are derived from CIFAR-100, which comprises 100 classes and 600 images per class, respectively. %Unlike the previous datasets, 
CIFAR-FS and FC100 are characterized by small-resolution images, with dimensions of 32 × 32 pixels. Specifically, CIFAR-FS is randomly divided into 64 training classes, 16 validation classes, and 20 testing classes. On the other hand, FC100 contains 100 classes sourced from 36 super-classes in CIFAR100. These 36 super-classes are further divided into 12 training super-classes (containing 60 classes), 4 validation super-classes (containing 20 classes), and 4 testing super-classes (containing 20 classes).

%For fine-grained classification, we utilize the CUB-200-2011 (CUB) dataset~\cite{cub}. 
The CUB-200-2011 (CUB) dataset is for fine-grained classification. 
It is composed of 200 classes with a total of 11,788 images. %To split the dataset, 
We follow %the protocol introduced by
the original setting~\cite{cub}, that randomly allocates 100 classes as the base set, 50 classes for validation, and 50 classes for testing.

\subsection{Implementation Details}
Motivated by the scalability and effectiveness of pretraining techniques, we employ an MIM-pretrained Vision Transformer as our model backbone. Specifically, we utilize the ViT-based architecture with a patch size of 16.
In the pre-training phase, we employ the same strategy proposed by \cite{FewTrue} to pretrain our ViT-Small backbones and mostly stick to the hyperparameter settings reported in their work. For the ViT-Base model, we employ the MAE-pretrained~\cite{MAE} model directly as our pre-trained model.
In the meta training phase, we utilize the AdamW optimizer with default settings. The initial learning rate is set to $1e-5$ and decays to $1e-6$ following a cosine learning rate scheme. %schedule. 
For the $mini$ImageNet and $tiered$ImageNet datasets, we resize the images to a size of 224 and train for 100 epochs, with each epoch consisting of 600 episodes. Similarly, for the CIFAR-FS and FC100 datasets, we resize the images to a size of 224 and train for 90 epochs, with 600 episodes each. During the fine-tuning, we focus on the layers before and after our intra-task mutual attention method, as well as the CLS token. To be consistent with prior studies, we employ the validation set to select the best performing models.
In addition, we apply standard data augmentation techniques, including random resized loops and horizontal flips.

During the meta-testing phase, we randomly sample 1000 tasks, each containing 10 query images per class. We report the mean accuracy along with the corresponding $95\%$ confidence interval.

\subsection{Performance Comparison}

\begin{table*}[ht]
\caption{Comparison with the state-of-the-art 5-way 1-shot and 5-way 5-shot performance with 95$\%$ confidence intervals on fine-grained dataset CUB. 
% ViT-Base is the backbone, with parameters of the last two layers plus the CLS token be fine-tuned. The top two results are highlighted in bold and italic.
} % based on their relative rankings.}
\label{tab:cub} 
\begin{center}
% \scalebox{1.0}{
    \begin{tabular}{l|cr|cc}
    \hline
    % \cline{1-1} \cline{3-5}
    \multirow{2}{*}{Model} &\multirow{2}{*}{Backbone} & \multirow{2}{*}{$\approx$Params} & \multicolumn{2}{c}{CUB}  \\
    \multicolumn{1}{l|}{} & \multicolumn{1}{l}{}& \multicolumn{1}{l|}{} & 1-shot & 5-shot            \\ 
    \hline
    \hline
    ProtoNet~\cite{PrototypicalNetwork} & Conv-4 & 0.1 M& 64.42±0.48 & 81.82±0.35 \\
    FEAT~\cite{FEAT} & Conv-4 & 0.1 M& 68.87±0.22 & 82.90±0.15 \\
    MELR~\cite{MELR} & Conv-4 & 0.1 M& 70.26±0.50 & 85.01±0.32 \\
    HGNN~\cite{hgnn} & Conv-4 & 0.1 M& 69.43±0.49 &87.67±0.45 \\
    \hline
    % MVT~\cite{MVT} & ResNet-10 & 4.9 M& – & 85.35±0.55 \\
    MatchNet~\cite{MatchingNetwork} & ResNet-12 & 12.4 M& 71.87±0.85 & 85.08±0.57 \\
    ICI~\cite{ICI} & ResNet-12 & 12.4 M& 76.16 & 90.32 \\
    RENet~\cite{RENet} &ResNet-12 & 12.4 M&79.49±0.44 &91.11±0.24 \\
    MAML~\cite{MAML} & ResNet-18 & 11.2 M& 68.42±1.07 & 83.47±0.62 \\
    % $\Delta$-encoder~\cite{delta-encoder} & ResNet-18 & 11.2 M& 69.80 & 82.60 \\
    Baseline++~\cite{baseline++} & ResNet-18 & 11.2 M& 67.02±0.90 & 83.58±0.54 \\
    AA~\cite{AA} & ResNet-18 & 11.2 M& 74.22±1.09 & 88.65±0.55 \\
    Neg-Cosine~\cite{Neg-cosine} & ResNet-18 & 11.2 M& 72.66±0.85 & 89.40±0.43 \\
    % LaplacianShot~\cite{Laplacianshot} & ResNet-18 & 11.2 M& 80.96 & 88.68 \\
    Meta DeepBDC~\cite{Meta-DeepBDC} & ResNet-18 & 11.2 M& 83.55±0.40 & 93.82±0.17 \\
    STL DeepBDC~\cite{Meta-DeepBDC} & ResNet-18 & 11.2 M&  84.01±0.42 &  94.02±0.24 \\
    \hline
    % IMAformer (ours) & ViT-Small & 22 M& \bf 85.02±0.51	& \bf 94.26±0.45 \\
    IMAformer (ours) & ViT-Small & 22 M& \bf 89.30±0.29	& \bf 95.80±0.18 \\
    % \hline
    % \hline
    IMAformer (ours) & ViT-Base & 86 M& 86.57±0.56	& 95.25±0.35 \\
    \hline
    \end{tabular}
    % }
\end{center}

\end{table*}

Following the established conventions of few-shot learning, we perform experiments on five popular few-shot classification benchmarks, and the results are shown in Tables~\ref{tab:miniImage},~\ref{tab:cifar} and~\ref{tab:cub}, respectively. 
From the results, it is evident that our proposed IMAformer consistently achieves competitive performance compared to the state-of-the-art methods on both the 5-way 1-shot and 5-way 5-shot tasks.

\textbf{Results on $mini$Imagenet and $tiered$Imagenet datasets}. Table \ref{tab:miniImage} presents a comparison on the 1-shot and 5-shot performance of our method with the baselines on the $mini$Imagenet and $tiered$Imagenet datasets. We achieve significant improvements over existing  SOTA results while utilizing fewer learnable parameters. For instance, on the $mini$Imagenet dataset, with ViT-Small as the backbone, our IMAformer surpasses the runner-up by 3.74\% in the 1-shot setting and 1.80\% in the 5-shot setting. For ViT-Base backbone, our IMAformer surpasses the runner-up by 10.94\% in the 1-shot setting and 6.90\% in the 5-shot setting. Similarly, on the $tiered$Imagenet dataset, with ViT-Small as the backbone, IMAformer outperforms the runner-up  by 0.12\% and 1.05\% in the 1-shot and 5-shot settings, respectively. For ViT-Base backbone, IMAformer outperforms the runner-up  by 3.43\% and 3.14\% in the 1-shot and 5-shot settings, respectively.

Compared to FewTURE and HCTransformer, which require external module, our method eliminates this need while achieving better performance and utilizing fewer learnable parameters. The significant margin between our method and the runner-up baselines further validates the contribution of our network structure, which effectively captures inherent information in the data and maintains good generalization capability.

\textbf{Results on small-resolution datasets}. To verify the adaptability of the model, we conduct further experiments on two small-resolution datasets along with comparisons to other methods. This allows us to evaluate the performance of our method across various data scenarios, ensuring fair and comprehensive performance analysis.

Table~\ref{tab:cifar} demonstrates the 1-shot and 5-shot classification performance on the small-resolution datasets CIFAR-FS and FC100. In the case of CIFAR-FS, IMAformer outperforms the 
runner-up by 2.57\% in the 1-shot setting and 3.16\% in the 5-shot setting. Similarly, for FC100, our method surpasses the runner-up  by 1.77\% in the 5-shot setting.

\textbf{Results on fine-grained dataset}.
Table~\ref{tab:cub} showcases the 1-shot and 5-shot classification results on the fine-grained dataset CUB. 
Our IMAformer achieves improvements of 5.29\% in the 1-shot setting and 1.78\% in the 5-shot setting over the baselines with ViT-Small as the backbone. 

These results highlight the effectiveness of our method in both small-resolution and fine-grained classification scenarios. Moreover, it can be found that our method is effective regardless of whether it adopts ViT-Small or ViT-Base. Our method achieves state-of-the-art performance across all settings, with only one exception on the 1-shot task on FC100. In this particular scenario, the performance of our method is slightly lower compared to HCTransformer~\cite{HCTransformer}, which utilizes a stacked Transformers architecture with a deeper backbone as the feature extractor.

% These significant improvements reinforce the superiority of our method, which leverages the intra-task mutual attention module to enhance the differentiation between support and query sets. The effectiveness of this approach is evidenced by the notable performance gains achieved in various few-shot learning tasks.
The substantial enhancements observed affirm the superiority of our method, which employs the intra-task mutual attention method to augment the distinction between support and query sets. The effectiveness of this approach is substantiated by the remarkable performance gains achieved across diverse few-shot learning tasks.

\subsection{Ablation Study}
In this subsection, we conduct ablation experiments to analyze the impact of each component on the performance of our method. We focus on the 5-way 5-shot setting on the $mini$Imagenet dataset, utilizing the pre-trained Vision Transformer backbone. By analyzing the performance changes resulting from these ablations, we gain insights into the contribution of each component to the overall effectiveness of our method. Specifically, we investigate the influence of the following two components:

\textbf{Intra-task Mutual Attention Method}: We evaluate the performance of our model without the intra-task mutual attention method, which is responsible for enhancing the interaction and alignment between the support and query sets.

\begin{table}[ht]
\caption{Comparison on 5-way 5-shot performance with 95$\%$ confidence intervals 
with the model without intra-task mutual attention method 
on $mini$ImageNet. ViT-Base is the backbone, with parameters of the last two layers and the CLS token be fine-tuned.}
\label{tab:cait}
\begin{center}
    \begin{tabular}{l|ccc}
    \hline
    % \cline{1-1} \cline{3-5}
    Model & Backbone & Fine-tuning Layers& Accuracy \\ 
    \hline
    \hline
    Vanilla & ViT-Small & no layer & 82.05±0.34\\
    Vanilla & ViT-Small & last 6 layers & 75.50±0.51\\
    Vanilla & ViT-Small & 12 layers & 75.16±0.51\\
    IMAformer & ViT-Small & last 6 layers & 88.18±0.29\\
    \hline
    % Vanilla & ViT-Base & no layer & 39.69±0.74\\
    Vanilla & ViT-Base & last 2 layers+CLS & 75.68±0.81\\
    IMAformer & ViT-Base & last 2 layers+CLS & 93.28±0.42\\
    \hline
    \end{tabular}
\end{center}

\end{table}

To verify the effectiveness of our intra-task mutual attention method, we design a comparative %experiment
model called Vanilla. In the Vanilla method, we do not employ patch token switching, and the final CLS tokens of the support and query sets are solely encoded by the backbone with the last six layers be fine-tuned. 
As shown in Table~\ref{tab:cait}, the results clearly demonstrate the significant improvement achieved by our IMAformer method compared to the Vanilla method no matter on ViT-Small or ViT-Base. 

Furthermore, an observation from the initial three rows in Table~\ref{tab:cait} indicates that the absence of our intra-task mutual attention method leads to a degradation in the performance of the fine-tuned model. 
This result confirms the efficacy of our intra-task mutual attention method in improving the performance of few-shot learning tasks.

\begin{figure}[t] %插入图片
\centering %图片居中

\begin{tikzpicture}[scale=0.85] %tikz图片

\begin{axis}[
    xlabel=Fine-tuning layers, %横坐标名
    % \vspace{3mm}
    ylabel=Accuracy on test set $(\%)$, %纵坐标名
    tick align=inside, %刻度在外显式
    legend style={at={(0.80,0.35))},anchor=north} %图例在图下方显示
    ]

%第一条线，mark是折线标示形状
\addplot[smooth,mark=square,blue, line width = 1.5pt] plot coordinates { 
    (1,86.11)
    (2,86.5)
    (3,87.35)
    (4,87.82)
    (5,88.05)
    (6,88.18)
    (7,88.13) 
    (8,88.28)
    (9,88.17)
    (10,88.23)
    (11,88.26)
    (12,88.18)

};
\addlegendentry{w/o CLS}

\addplot[smooth,mark=*,red, line width = 1.5pt] plot coordinates { 
    (1,85.80)
    (2,86.21)
    (3,87.35)
    (4,87.71)
    (5,88.01)
    (6,88.18)
    (7,88.13)
    (8,88.25)
    (9,88.15)
    (10,88.14)
    (11,88.25)
    (12,88.19)

};
\addlegendentry{w/ CLS}
%第二条线，mark是折线标示形状

\end{axis}
\end{tikzpicture}
\caption{Comparison on 5-way 5-shot performance with the model under different fine-tuning layers on $mini$ImageNet. ViT-Small is the backbone. 'w/o CLS' denotes fine-tuning excluding CLS and 'w/ CLS' denotes fine-tuning including CLS.}
\label{fig:fine-tune}
\end{figure}
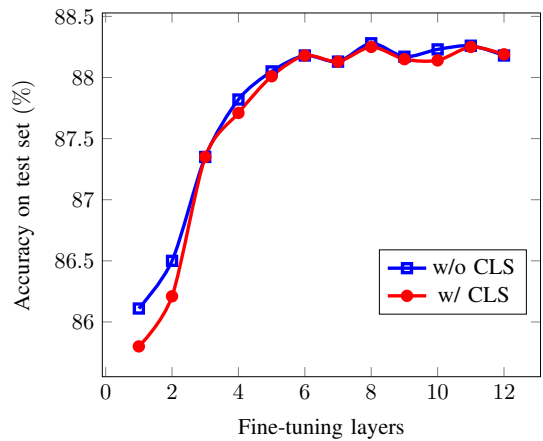

\textbf{CLS Token and Learnable Parameters}: We conduct an examination of the impact of various fine-tuning parameters on the performance of the model, including the fine-tuning of the CLS token.

We conducted a more in-depth analysis to assess the influence of various fine-tuning parameters on our method, utilizing the ViT-Small architecture as the basis. The corresponding results are presented in Figure~\ref{fig:fine-tune}. Notably, as the number of fine-tuning layers increases, there is a continuous improvement in performance. However, beyond six fine-tuning layers, the model experiences slight fluctuations. Further increasing the number of fine-tuning layers results in an enhancement in model performance, but the improvement is not substantial.
Additionally, we observe that not fine-tuning the CLS token yields higher accuracy rates than fine-tuning. 
%Ultimately, we decide to proceed with fine-tuning the last two layers and the CLS token.
Thereby, for ViT-Small, we choose to fine-tune the last six layers. 
% \HK{Why there is a clear decay for 7 or 8 layers?}

\begin{table}[t]
\caption{Comparison on 5-way 5-shot performance with 95$\%$ confidence intervals
with the model under different fine-tuning parameters
on $mini$ImageNet. ViT-Base is the backbone.}
\label{tab:fine-tune}
\begin{center}
    \begin{tabular}{c|lcc}
    \hline
    % \cline{1-1} \cline{3-5}
    Num & Fine-tuning Layers & $\approx$Params & Accuracy \\ 
    \hline
    \hline
    1 & no layer & ~~0 M & 38.82±0.74\\
    2 & last 1 layer & ~~7 M &  90.71±0.48\\
    3 & last 2 layers & 14 M & 92.66±0.44\\
    4 & last 2 layers+CLS & 14 M & \bf 93.28±0.42\\
    5 & last 3 layers+CLS & 21 M & 92.83±0.45\\
    6 & last 4 layers & 28 M & 86.48±0.58\\
    7 & last 4 layers+CLS & 28 M & 87.85±0.55\\
    \hline
    \end{tabular}
\end{center}
\end{table}

Simultaneously, we conducted an analysis to assess the impact of different fine-tuning parameters on our approach based on ViT-Base. The corresponding results are presented in Table~\ref{tab:fine-tune}. 
Notably, when fine-tuning solely one layer, the performance of the model exhibited improvements compared to the instance where no fine-tuning is applied. Moreover, as the number of learnable parameters increases, the model's performance  improves correspondingly.
However,  
note that there exists a trade-off between performance and the number of learnable parameters. Specifically, the best performance was achieved when fine-tuning the last two layers together with CLS token; beyond this point, the accuracy began to decline as more parameters were fine-tuned. Additionally, we observe that fine-tuning the CLS token yields higher accuracy rates than no-fine-tuning.

\subsection{Visualization Analysis}
\begin{figure}[t!]
\centering
\subfloat[]{\includegraphics[width=0.49\linewidth]{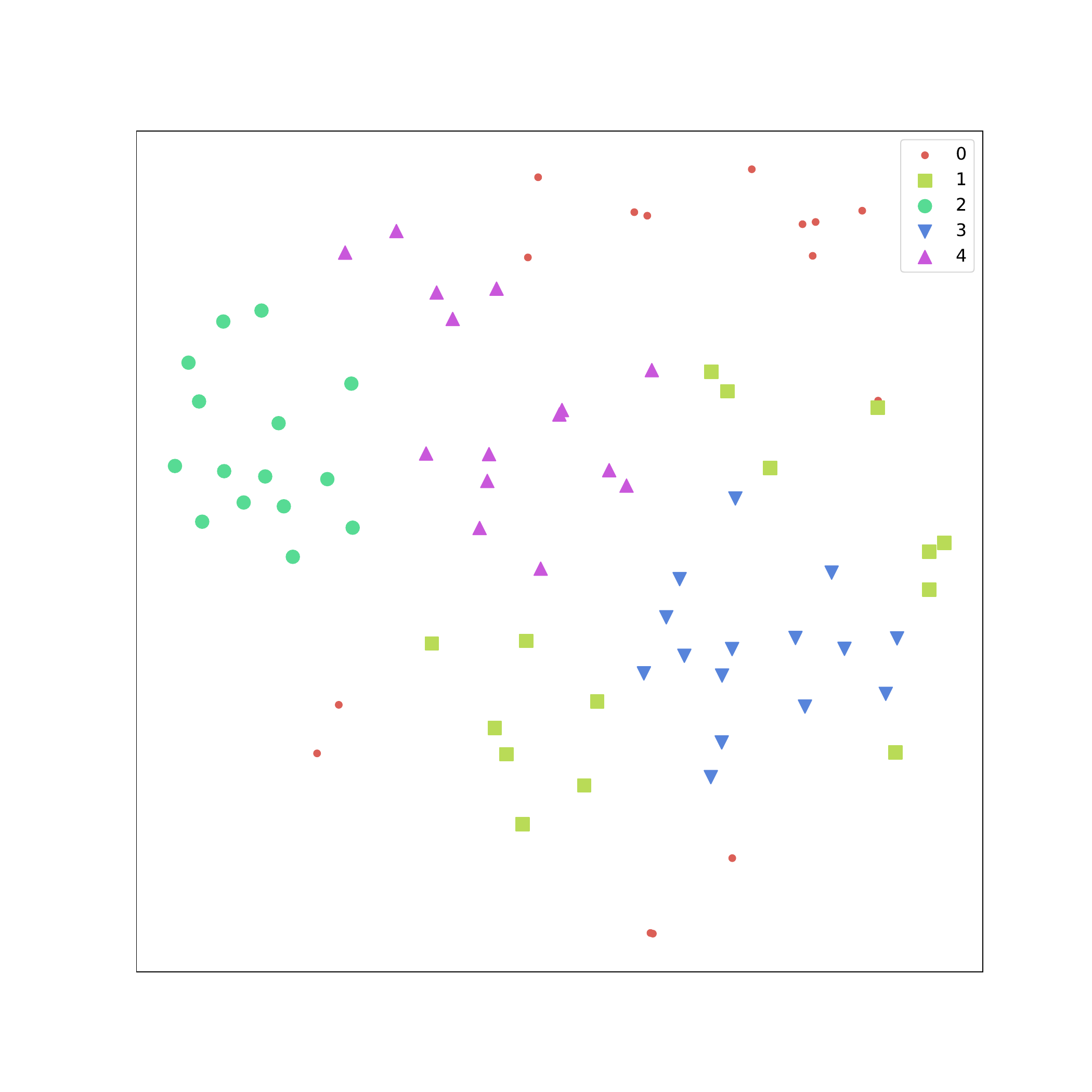}%
\label{fig1}}
\hfil
\subfloat[]{\includegraphics[width=0.49\linewidth]{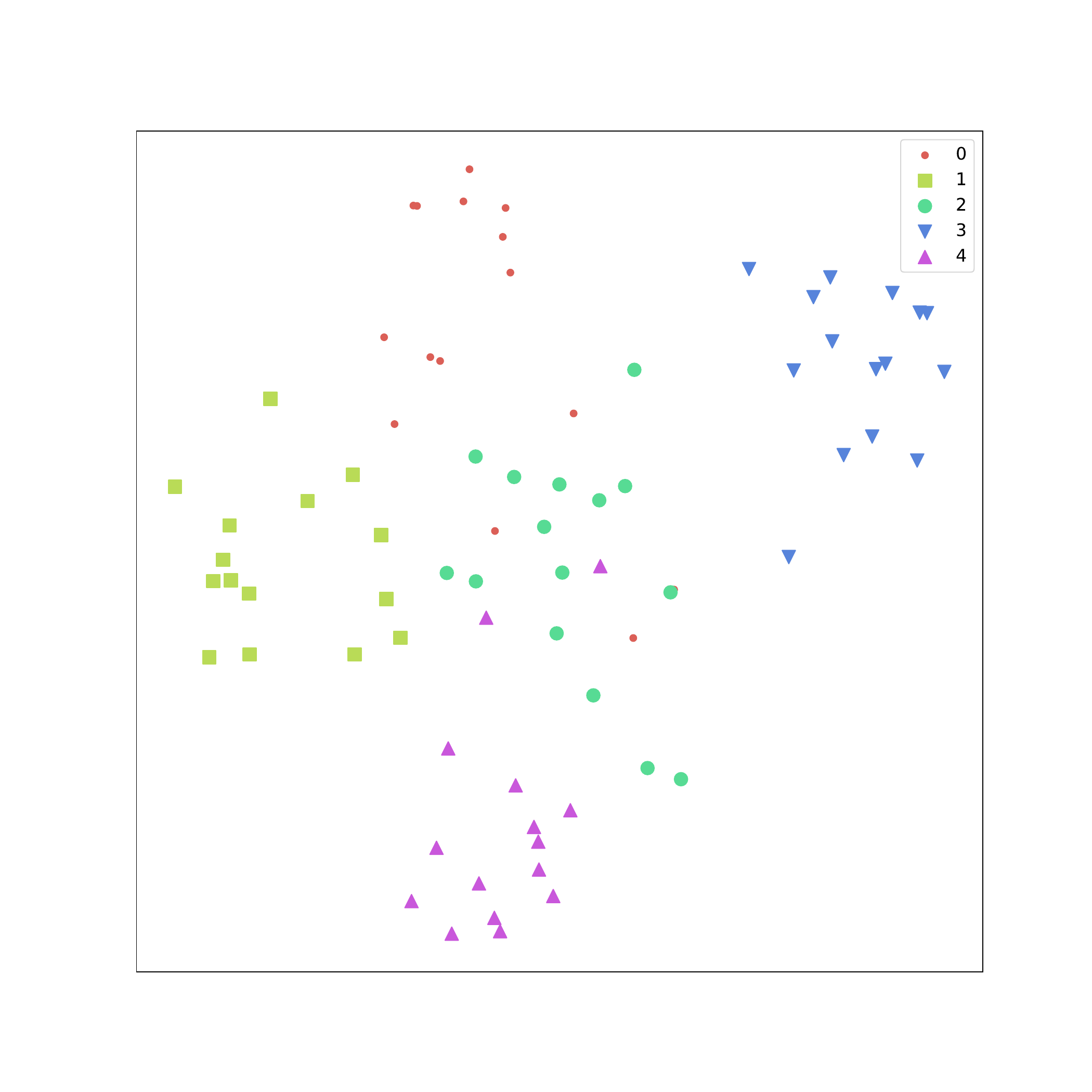}%
\label{fig2}}
\caption{Qualitative visualization of model-based embedding before and after using intra-task mutual attention method on test tasks. Each figure shows the locations of PCA projected query embeddings (a) before and (b) after the adaptation of IMAformer. Values below are the 5-way 15-shot few-shot task before and after the adaptation. Obviously, the embedding adaptation step of IMAformer pushes the query embeddings toward their own clusters, such that they can better fits the test data of its categories. }
\label{fig:tsne}
\end{figure}

To qualitatively evaluate the proposed intra-task mutual attention mechanism, we compare the query embedding visualization results of IMAformer to the model without intra-task mutual attention method. As shown in Fig.  \ref{fig:tsne}, after the mutual attention within the task, similar query features within the task are more clustered such that the visual features are most discriminative for a given task. 

This result once again demonstrates the effectiveness of our approach, that is, through the mutual attention of global features and local features within the task, the obtained query features and support features are more  distinguishable.

\section{Conclusion}
We proposed a novel framework called IMAFormer %to address the challenge of utilizing
that fully utilizes the global feature and local feature in the context of few-shot learning. Our approach involves segmenting the input image into smaller patches and encoding these local patches using a pre-trained Vision Transformer architecture, achieved through self-supervised learning. This segmentation allows us to obtain both global information representation through the CLS token and local information representation through the patch tokens. 
To enhance the representations of support and query sets, we introduced an intra-task mutual attention method. This method enables the support set focus on and leverage the detailed features presented in the query set, thereby enhancing its representations and promoting similarity between instances. Similarly, the query set can attend to the detailed features of each class in the support set, leveraging similar features to improve its representations and making the class features more representative. 
Empirical evaluations demonstrated the effectiveness of our proposed method,  achieving advanced %state-of-the-art 
performance. % on widely used few-shot classification datasets. 

Our approach offers several advantages. Firstly, it significantly improves the image representation through integrating class token and patch tokens. Secondly, it allows for efficient utilization of the knowledge and representations encoded in these pre-trained models learned from Masked Image Modeling pre-training task. By building upon these pre-trained models, we can potentially achieve competitive performance in few-shot learning tasks without any external weight module.

\section*{Acknowledgments}
This work is supported by National Natural Science Foundation (U22B2017).

%------------------------------------------------------------------------------------------------------------
%---------------------------------------------------------------------------------------------------------------------------

% \section*{Acknowledgments}
% This should be a simple paragraph before the References to thank those individuals and institutions who have supported your work on this article.

% {\appendix[Proof of the Zonklar Equations]
% Use $\backslash${\tt{appendix}} if you have a single appendix:
% Do not use $\backslash${\tt{section}} anymore after $\backslash${\tt{appendix}}, only $\backslash${\tt{section*}}.
% If you have multiple appendixes use $\backslash${\tt{appendices}} then use $\backslash${\tt{section}} to start each appendix.
% You must declare a $\backslash${\tt{section}} before using any $\backslash${\tt{subsection}} or using $\backslash${\tt{label}} ($\backslash${\tt{appendices}} by itself
%  starts a section numbered zero.)}


%-------------------references--------------------
\bibliographystyle{IEEEtran}
\bibliography{references}

\newpage

 

% \end{IEEEbiography}



\begin{IEEEbiography}[{\includegraphics[width=1in,height=1.25in,clip,keepaspectratio]{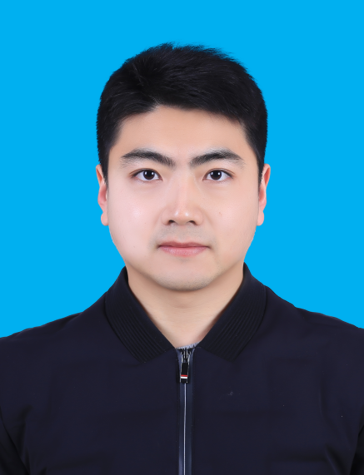}}]{Weihao Jiang}
is currently pursuing a Ph.D. in School of Computer Science and Technology, Huazhong University of Science and Technology, Wuhan, China.
His research focuses on deep learning,
few-shot learning and meta-learning.
\end{IEEEbiography}

\vspace{11pt}

\begin{IEEEbiography}[{\includegraphics[width=1in,height=1.25in,clip]{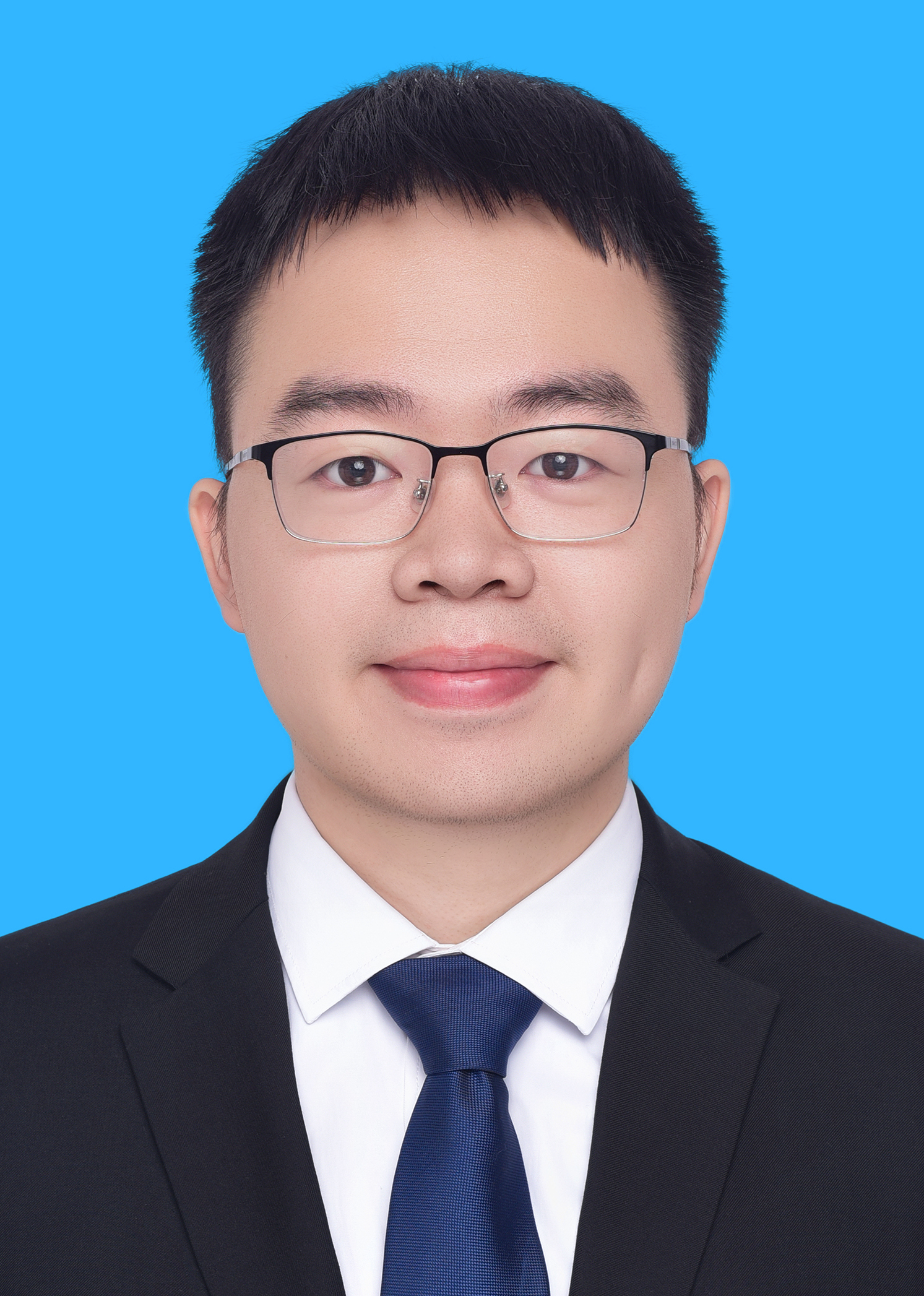}}]{Chang Liu}
is currently pursuing a master degree in computer application technology from the Huazhong University of Science and Technology, Wuhan, China. 
His research mainly focuses on deep learning, few-shot learning, and data augmentation.
\end{IEEEbiography}

\vspace{11pt}

\begin{IEEEbiography}[{\includegraphics[width=1in,height=1.25in,clip,keepaspectratio]{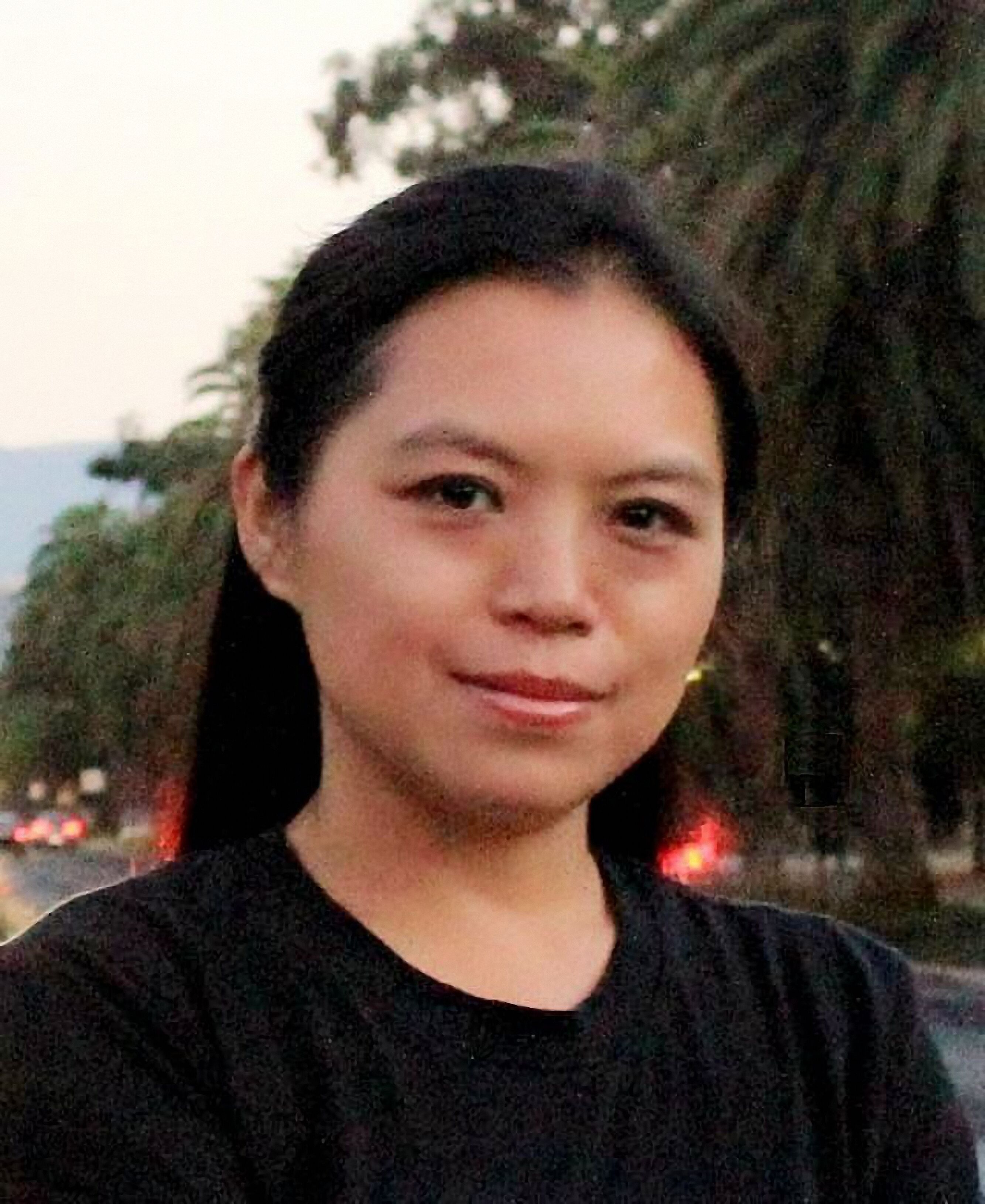}}]{Kun He}
(SM18) is currently a Professor in School of Computer Science and Technology, Huazhong University of Science and Technology, Wuhan, P.R. China. She received the Ph.D. degree in system engineering from Huazhong University of Science and Technology, Wuhan, China, in 2006. She had been with the Department of Management Science and Engineering at Stanford University in 2011-2012 as a visiting researcher. She had been with the de- partment of Computer Science at Cornell University in 2013-2015 as a visiting associate professor, in 
2016 as a visiting professor, and in 2018 as a visiting professor. She was honored as a Mary Shepard B. Upson visiting professor for the 2016-2017 Academic year in Engineering, Cornell University, New York. Her research interests include adversarial learning, representation learning, social network analysis, and combinatorial optimization. 
\end{IEEEbiography}

\vfill

\end{document}